\documentclass{article}
 \pdfoutput=1

\usepackage{arxiv}
\usepackage[utf8]{inputenc} 
\usepackage[T1]{fontenc}    
\usepackage{hyperref}       
\usepackage{url}            
\usepackage{booktabs}       
\usepackage{amsfonts}       
\usepackage{nicefrac}       
\usepackage{microtype}      
\usepackage{lipsum}		
\usepackage{graphicx}
\usepackage{natbib}
\setcitestyle{numbers,square}
\usepackage{doi}
\usepackage{makecell}
\usepackage{multirow}
\usepackage{color}
\usepackage[ruled]{algorithm2e}
\usepackage{multicol}
\usepackage{bm}
\usepackage{caption}

\title{The 2nd Place Solution for 2023 Waymo Open Sim Agents Challenge}

\author{{\hspace{1mm}Cheng Qian\thanks{This work was done during internship at DiDi Autonomous Driving}} \\
	College of Information Science and Engineering\\
	Northeastern University\\
	Shenyang, China \\
	\texttt{qiancheng625@gmail.com} \\
	\And
	{\hspace{1mm}Di Xiu$^*$} \\
	School of Electronic, Electrical and Communication Engineering\\
	University of Chinese Academy of Sciences\\
	Beijing, China \\
	\texttt{suoliweng98@gmail.com} \\
        \And
	{\hspace{1mm}Minghao Tian} \\
	DiDi Autonomous Driving\\
	  Beijing, China \\
	\texttt{tianminghao@didiglobal.com} \\
}

\renewcommand{\shorttitle}{The 2nd Place Solution for 2023 WOSAC}

\hypersetup{
pdftitle={The 2nd Place Solution for 2023 Waymo Open Sim Agents Challenge},
pdfsubject={},
pdfauthor={Cheng Qian, Minghao Tian, Di Xiu},
}

\begin{document}
\maketitle

\begin{abstract}
In this technical report, we present the 2nd place solution of 2023 Waymo Open Sim Agents Challenge (WOSAC) \cite{montali2023waymo}. We propose a simple yet effective autoregressive method for simulating multi-agent behaviors, which is built upon a well-known multimodal motion forecasting framework called Motion Transformer (MTR) \citep{shi2022mtr} with postprocessing algorithms applied. Our submission named \textit{MTR+++} achieves 0.4697 on the Realism Meta metric in 2023 WOSAC. Besides, a modified model based on MTR named \textit{MTR\_E} is proposed after the challenge, which has a better score 0.4911 and is ranked the 3rd on the leaderboard of WOSAC as of June 25, 2023.
\end{abstract}


\section{Introduction}
How to validate the safety of an autonomous driving system (ADS) is an important topic in both research community and L2+ autonomous driving industry. In UNECE document `Proposal for the Future Certification of Automated/Autonomous Driving Systems' (ECE/TRANS/WP.29/GRVA/2019/13), the authors proposed the so-called three pillars of safety validation: simulation, closed-road scenario test and real-world road test. Among the three, simulation is undoubtedly the safest and the most cost-efficient one. However, simulating traffic participants in a realistic way can be very challenging. 

WOSAC \cite{montali2023waymo} is a new challenge focusing on realistically simulating traffic agents in an interactive way. Contestants are asked to generate the future $8s$ behaviors of the agents in an autoregressive pattern, given the last $1.1s$ historical motion data together with the map features of the scenario. The challenge is actually very similar to motion forecasting in literature \citep{chai2019multipath,pmlr-v205-luo23a,9812107}. Here, we adapt one of the methods called Motion Transformer (MTR)  \cite{shi2022mtr} to help us generate the possible futures.

\vspace{-0.08in}
\paragraph{Motion Transformer (MTR) framework.}
Motion Transformer (MTR) framework is a transformer encoder-decoder model which achieved the SOTA performance on benchmarks in the 2022 Waymo Open Dataset Challenges. It uses a vectorized representation to handle both input trajectories and road map as poly-lines, and adopts the agent-centric strategy that organizes all inputs to the local coordinate system centered at that agent. During inference, the input of the MTR model is the past $1.1s$ (10Hz sampling) of historical motion data $\left\{ [cx,cy,cz,dx,dy,dz,\theta, vel_x,vel_y,valid] \right\}_{i \in [11]}$ of all the traffic participants (the local map features are included by collecting the map poly-lines nearby for each agent), where $[cx,cy,cz]$ represent the location; the $dx,dy,dz,\theta$ are the length, width, height and heading, respectively; the $vel_x,vel_y$ are the $x,y$-direction velocity and the $valid$ is a dataset-related feature which represents whether the ground-truth provided is valid. The output of the model are the predicted trajectories in form of $\left\{ [cx, cy, vel_x,vel_y] \right\}_{j \in [80]}$. We run the same inference procedure for each agent in one inference round.

\section{Method}
\label{sec:method}
We propose a simple method to simulate the behaviors of the agents given the past historical motion data. MTR framework is used as the motion forecasting block to produce a hybrid of open-loop and closed-loop motion data by executing model inference autoregressively at 0.5Hz (more discussion on the update rate can be found in Section \ref{sec:experiment:update_rate}). One inference round includes 6 possible future trajectories generated for each agent (with sum-to-one probabilities assigned). At any simulation step, agents utilize MTR-generated trajectories and conduct the collision-mitigation policy (introduced in Section \ref{sec:method:policy}) to pick 1 out of 6. 
To be more specific, the MTR model, together with the collision-mitigation policy, inferring a future trajectory $\left\{ [cx, cy, vel_x,vel_y] \right\}_{j \in [20]}$ for each agent over a $2s$ horizon, based on the past $1.1s$ of historical motion data, along with map features. After applying a heading calculation (introduced in Section \ref{sec:method:heading}) to the trajectory, it will be executed as a motion plan, and the last $1.1s$ of it will be consumed by the MTR model as the past $1.1s$ history at the next simulation (update) step. During the entire simulation, we keep the $cz,dx,dy,dz$ of each agent as constants (directly read from historical motion data). After running several rounds, a simulated scenario will be generated. To generate 32 results as required by the challenge, we simply select 32 variants of the fine-tuned MTR models and each generates one. More details can be found in Section \ref{sec:experiments:details}. This method achieves 0.4697 on the Realism Meta metric, which is the 2rd place of 2023 WOSAC. 

\vspace{-0.08in}
\paragraph{MTR\_E --- A modified model based on MTR.}
We also propose a modified model based on MTR, called MTR\_E, which can directly predict the $\left\{ [cx,cy,cz,\theta,vel_x,vel_y] \right\}_{j \in [80]}$ of the agents. To be more specific, we simply modified the decoder part of MTR. We extended the output dimension of MLP in dense future prediction module and motion prediction module to support output of $cz$ and $\theta$, while adding extra $L1$ losses for $cz$ and $\theta$ to the total loss. This method achieves 0.4911 on the Realism Meta metric, ranking 3rd on the leaderboards of 2023 WOSAC as of June 25, 2023.

\subsection{Heading calculation}
\label{sec:method:heading}
\begin{figure*}[h]
	\centering
 \begin{tabular}{ccc}
     \includegraphics[width=2in]{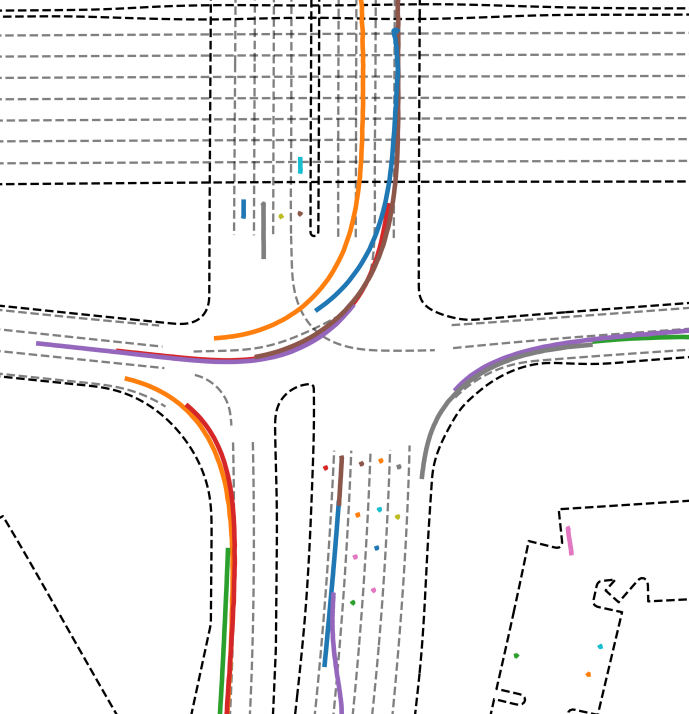} & \includegraphics[height=2in]{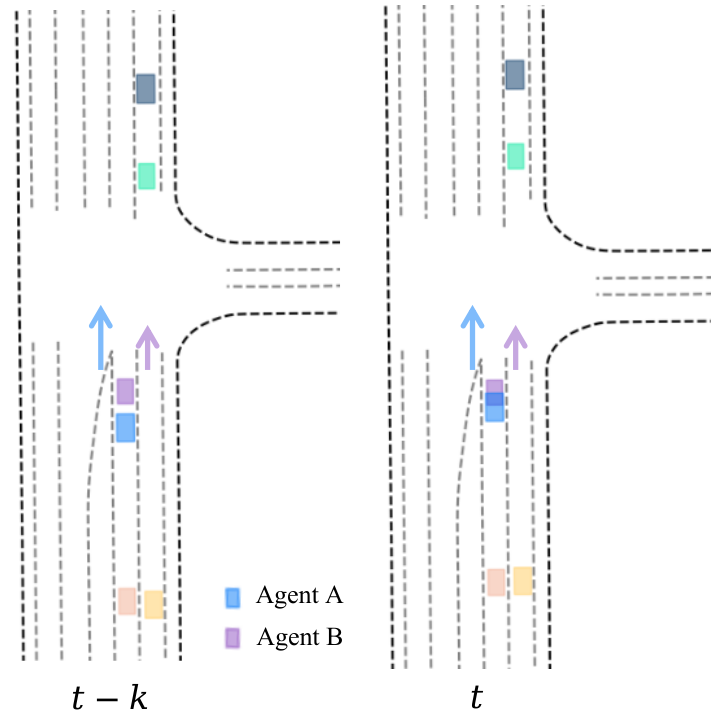} & \includegraphics[height=2in]{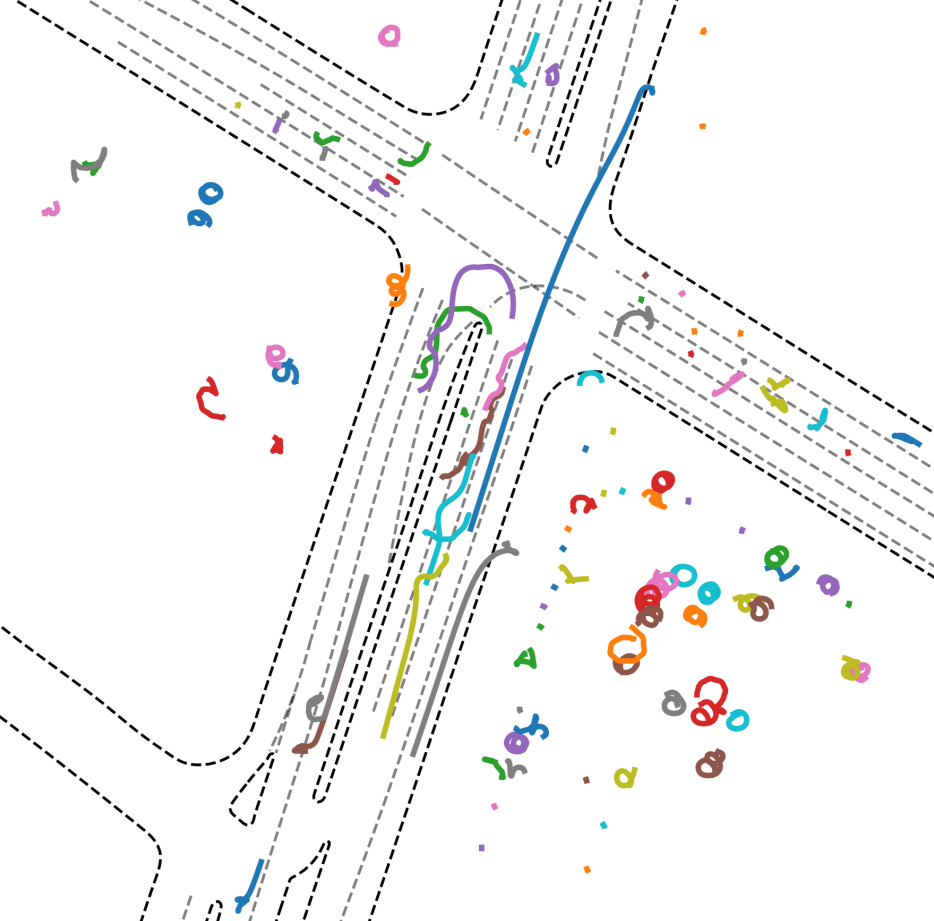} \\
     & & \\
     (a) & (b) & (c) 
 \end{tabular}
	
	\caption{(a) Trajectories directly output by MTR model; (b) Two vehicles collide after executing the trajectories with the highest probability scores as a motion plan; (c) Scenario generated at 10Hz;   }
	\label{fig:mtr_result}
\end{figure*}

Since MTR generates high-quality trajectories which is shown empirically in previous work (Figure \ref{fig:mtr_result}(a)), we simply use the $(x, y)$ coordinates of the generated trajectories to compute the corresponding headings of the agents (implemented using \textit{np.arctan2} to calculate the heading). We treat any agent with moving distance less than 0.3m during 2s according to its predicted trajectory as a stopped agent. For stopped agents, we keep their heading still. For those agents which exhibit extreme angular rotation rates, we stabilize the headings by checking the consecutive ones. If the difference between two consecutive headings $[t, t+0.1s]$ is larger than 0.3rad, we overwrite the heading at $t+0.1s$ and force it to be equal to the heading at $t$. Finally, we normalize $h_i$ such that $h_i \in [-\pi, \pi)$. 

\subsection{Collision-mitigation policy}
\label{sec:method:policy}
Suppose there are $N$ agents (including the ADV agent) to be simulated in a given scene. With 6 trajectories generated for each agent by MTR, there could be $6^N$ possible results at each update step. We observed that simply picking the trajectories with the largest probability could lead to scenarios with unrealistic collisions (Figure \ref{fig:mtr_result}(b)). 

To mitigate the presence of collisions while avoiding brute-force searching, we propose a greedy strategy to try to select one combination with reasonable small collision counts. The idea is to first build the $6N$ by $6N$ distance matrix $D$ where entry $D_{6m + i - 1, 6n + j - 1}$ indicates the minimum $L^2$-distance between the $i$th most likely trajectory of agent $m$ and the $j$th of agent $n$, where $0 \leq m,n \leq N - 1$ and $1 \leq i, j \leq 6$. If $D_{6m + i - 1, 6n + j - 1} \leq \left( {width}_m + {width}_n\right) / 2$ where ${width}_k$ is the width of agent $k$, then executing these two trajectories as a motion plan will lead to a collision. Thus, we can build a 0-1 matrix $C$ where $C_{6m + i - 1, 6n + j - 1} = 0$ if a collision happens between the $i$th most likely trajectory of agent $m$ and the $j$th of agent $n$, otherwise it will be set to $1$. Since trajectories generated for the same agent are all originated from the same starting position, these trajectories collide and thus we have $N$ $6 \times 6$ zero-blocks lying on the diagonal of $C$. The matrix $C$ is symmetric and can be viewed as an adjacency matrix of an undirected graph with $6N$ nodes (keep the same indexing of the matrix). We then run Algorithm \ref{alg:clique_finding} to find a dense subgraph of size $N$. Under the adjacency matrix setting, the density of a subgraph of size $l$ can be easily calculated by the sum of all the entries of the corresponding submatrix divided by $l \times (l - 1)$. A clique is a subgraph with density equal to $1$. Here, we say a subgraph is dense if its density is $\geq 0.95$. The denser the subgraph is, the less the collision count will be. Finally, we select the corresponding trajectories according to the dense subgraph, i.e. if vertex with index $(6m+i -1)$ is selected, then we pick the $i$th most likely trajectory of agent $m$.

\begin{algorithm}[h]
\caption{Clique/dense subgraph-finding heuristic}\label{alg:clique_finding}
\textbf{Input}:  $C$ --- the ($6N \times 6N$)-adjacency matrix described in Section \ref{sec:method:policy}\;
\textbf{Output}: $\bm{c}$ --- an array of indices of size $N$ indicating the vertices of the derived dense subgraph\;
1. Initialize $\bm{c} := \left[0, 6, \cdots, 6(N-1)\right]$\;
2. Check whether $\{\bm{c}[0],\cdots, \bm{c}[N-1]\}$ forms a clique; if so, stop and OUTPUT $\bm{c}$\;
3. Compute $deg$ the degree array of vertices\;
4. Define two recursive functions $FindDSG(i, l, s)$ and $FindClique(i, l, s)$ which gradually expands the subgraph. Here, $i$ indicates the starting index of the search for the $l$th joiner vertex, expecting to find one with degree at least $(s-1)$:\\
DEF $FindDSG(i, l, s):$\\
~~~~FOR $j \in [i, i+5]$:\\
~~~~~~~~IF $deg[j] \geq s - 1$:\\
~~~~~~~~~~~~SET $\bm{c}[l] = j$\;
~~~~~~~~~~~~IF $\{\bm{c}[0],\cdots, \bm{c}[l]\}$ forms a dense subgraph:\\
~~~~~~~~~~~~~~~~IF $l < s - 1$:\\
~~~~~~~~~~~~~~~~~~~~RUN $FindDSG \left(6 \times \left(l+1\right), l+1, s \right)$\;
~~~~~~~~~~~~~~~~ELSE stop and OUTPUT $\bm{c}$\;
~~\\
DEF $FindClique(i, l, s):$\\
~~~~LET $j_{tmp} := -1$\;
~~~~FOR $j \in [i, i+5]$:\\
~~~~~~~~SET $\bm{c}[l] := j$\;
~~~~~~~~IF $\{\bm{c}[0],\cdots, \bm{c}[l]\}$ forms a clique:\\
~~~~~~~~~~~~SET $j_{tmp} := j$ and BREAK\;
~~~~IF $j_{tmp} \neq -1$:\\
~~~~~~~~SET $\bm{c}[l] := j_{tmp}$\;
~~~~~~~~IF $l < s - 1$:\\
~~~~~~~~~~~~RUN $FindClique \left(6 \times \left(l+1\right), l+1, s \right)$\;
~~~~~~~~ELSE stop and OUTPUT $\bm{c}$\;
~~~~ELSE run $FindDSG(i, l, s)$\;
~~\\        
5. RUN $FindClique(0, 0, N)$
\end{algorithm}

\begin{table*}[t]
	\caption{Different update rate.}
	\centering
	\begin{tabular}{c|cccc|c}
 \toprule
		\makecell[c]{Update rate\\
  (repeated 32 times)}&\makecell[c]{\textbf{Realism Meta}\\\textbf{metric $\uparrow$}} & \makecell[c]{Kinematic\\metrics $\uparrow$} & \makecell[c]{Interactive\\metrics $\uparrow$} & \makecell[c]{Map-based\\metrics $\uparrow$} & minADE $\downarrow$ \\
		\midrule 
  10Hz & 0.2836 & 0.0934 & 0.1243 & 0.0659  & 8.9345 \\
  0.5Hz & \textbf{0.4311} & \textbf{0.1232} & \textbf{0.1554} & \textbf{0.1525} & \textbf{2.2093} \\
		\bottomrule
	\end{tabular}
	\label{table:hz}
\end{table*}

\begin{table*}[t]
	\caption{Effects of collision-mitigation policy.}
	\centering
	\begin{tabular}{c|cccc|c}
 \toprule
	\makecell[c]{Collision-mitigation\\policy}&\makecell[c]{\textbf{Realism Meta}\\\textbf{metric $\uparrow$}} & \makecell[c]{Kinematic\\metrics $\uparrow$} & \makecell[c]{Interactive\\metrics $\uparrow$} & \makecell[c]{Map-based\\metrics $\uparrow$} & minADE $\downarrow$ \\
		\midrule 
$\times$  & 0.4743 & 0.1298 & 0.1747 & 0.1698  & 1.7560  \\
   $\checkmark$   & \textbf{0.4753} & \textbf{0.1299} & \textbf{0.1754} & \textbf{0.1700}  & \textbf{1.7554}  
  \\ 
		\bottomrule
	\end{tabular}
	\label{table:ablation}
\end{table*}

\begin{table*}[t]
\centering
\caption{Effects of number of variants of MTR model selected.}
 	\begin{tabular}{c|cccc|c}
 \toprule
		\makecell[c]{number of variants} &  \makecell[c]{\textbf{Realism Meta} \\\textbf{metric $\uparrow$}} & \makecell[c]{Kinematic \\metrics $\uparrow$} & \makecell[c]{Interactive \\metrics $\uparrow$} & \makecell[c]{Map-based \\ metrics $\uparrow$} & minADE $\downarrow$ \\
		\midrule
		 1 (repeat 32 times) & 0.4336 & 0.1228 & 0.1580 & 0.1528 & 2.1848\\
   1 [1,2,4,4]  & 0.4398 & 0.1194 & 0.1593 & 0.1611  & 2.1674  \\
		2 [1,1,4,4]  & 0.4487 & 0.1209 & 0.1646 & 0.1632 & 2.1373      \\
	4 [1,1,2,4]      & 0.4519 & 0.1221 & 0.1659 & 0.1639 & 2.0618  \\
 8 [1,1,1,4]      & 0.4652 & 0.1286 & 0.1690 & 0.1675 & 1.9015  \\
 16 [1,1,1,2]      & 0.4712 & 0.1288 & 0.1732 & 0.1692 & 1.8096  \\
 32 [1,1,1,1] (\textbf{MTR+++})& 0.4753 & 0.1299 & \textbf{0.1754} & 0.1700 & 1.7554\\
 \midrule
 32 [1,1,1,1] (\textbf{MTR\_E})     & \textbf{0.4958} & \textbf{0.1484} & 0.1738 & \textbf{0.1735} & \textbf{1.6642}  \\
		\bottomrule
	\end{tabular}
	\label{table:variants}
\end{table*}

\section{Experiments}
Considering that the number of scenarios in the Waymo Open Motion Dataset (WOMD) \cite{9709630} v1.2 validation set is as high as 44,097 scenarios, it would be very time consuming to select all validation sets for evaluation each time, so all the evaluations in this section were conducted only on the first TFRecord file, which consists of 287 validation scenarios.  

\vspace{-0.08in}
\paragraph{WOSAC Metrics.} \label{sec:experiment:metric}  
In order to better evaluate the 32 scenarios submitted by the contestants, the WOSAC score submissions by estimating densities from history-conditioned sim agent rollouts, then computing the negative log likelihood of the logged future under this density. The futures are represented with behavior-characterizing measures from three categories: \textit{motion, agent interactions, road / map adherence}\footnote{https://waymo.com/open/challenges/2023/sim-agents/}. The kinematic-based features is the agents' motion attribute, include linear speed, linear acceleration magnitude, angular speed, and angular acceleration magnitude. The interaction-based features represents the agents' interactions, including Time-to-collision (TTC) , distance to nearest object and modified Gilbert–Johnson–Keerthi (GJK) algorithm implementation. The map-based features shows how the agents interact with the road / map, including road departures and distance to road edge. The realism meta-metric is the primary metric used for ranking the methods, which is the sum of kinematic, interactive and map-based metrics. The minADE metric is also calculated but it is the secondary tie-breaking metric used for event of a tie. For more datails about the metrics, please refer to \cite{montali2023waymo}.

\subsection{Implementation Details}
\label{sec:experiments:details}
We trained the MTR model with the same hyper-parameter settings as in \cite{shi2022mtr} on both the training sets of WOMD v1.1 and v1.2 except for the learning rate. Two rounds of pre-training were conducted based on WOMD v1.1 to get two baseline models (each ran 30 epochs with 8 GPUs (NVIDIA RTX A6000) and batch size of 128 scenes): the learning rate is decayed by a factor of 0.5 every 2 epochs from epoch 20 and epoch 22, respectively. These two pretrained models were further fine-tuned on WOMD v1.2 for 20 epochs. Therefore, we obtained 40 separate models, one for each of 40 epochs as candidates. Finally, we picked 32 models out of 40 with the highest mAP score, and applied them to generate scenarios in the autoregressive fashion introduced in Section \ref{sec:method}. Using models from the results of different epochs can not only be viewed as an ensemble method, but also an implicit way of adding noise which can generate more diverse multi-agent behaviors. Experiments also showed that the more models we included in the ensemble, the higher the overall composite metric would be (see Section \ref{sec:experiments:number_variants} below).


\subsection{Update rate} \label{sec:experiment:update_rate}  
 We attempted to use a frequency of 10Hz for autoregressive simulation, which resulted in poor performance (Table \ref{table:hz}). After checking the visualization (Figure \ref{fig:mtr_result}(c)), we observed that several generated trajectories were unrealistic like wandering on the map without specific purpose. This may be because each simulation step would bring a certain disturbance error to the predicted trend of the previous simulation step. The more simulation steps we conduct, the greater the accumulated error will be.

\subsection{Effects of collision-mitigation policy}
\label{sec:experiments:ablation}
We compare our collision-mitigation update policy with the update policy which always selects the trajectories with the highest probability scores. Table \ref{table:ablation} shows that our proposed collision-mitigation policy slightly improves the quality of the simulated scenarios. Interestingly, it improves all the metrics listed a little bit. 

\subsection{Number of MTR variants}\label{sec:experiments:number_variants}
Since our model runs at 0.5Hz, we have 4 simulation steps in total in 8s simulation duration. We use notation $[m_1, m_2, m_3, m_4]$ to denote the number of different scenarios generated by our model at each step. For example, $[1,2,4,4]$ means that we have generated 2 different scenarios at the second simulation step by picking the top-2 combinations of trajectories in the sense of the joint probability. We tested on $1,2,4,8,16,32$ variants respectively with heading calculation module and collision-mitigation policy enabled. Empirical results (Table \ref{table:variants}) showed that 32 variants are indeed needed and it significantly improves the Interactive metrics. 


\subsection{MTR\_E}\label{sec:experiments:MTR_E}
We show the performance of MTR\_E model introduced in Section \ref{sec:method}. Table \ref{table:variants}  shows the MTR\_E improves all the metrics listed except the interactive metrics, which leads to a $4.3\%$ increase on Realism Meta metric than our submission MTR+++. Besides, we also observe that minADE correlates quantitatively with the ordinal ranking provided by the Realism Meta metric. It would be interesting to explore the relationship between these two metrics.

\subsection{Results on WOSAC leaderboard}
Consistent with the observations in Section \ref{sec:experiments:MTR_E}, Table \ref{table:LB} shows the MTR\_E improves all the metrics listed except the interactive metrics, which empirically indicates that testing on the first TFRecord file of WOMD v1.2 validation set can quickly provide a rough estimate of the performance of the method.

The detailed performance comparison between MTR+++ and MTR\_E on WOSAC test set are presented in Table \ref{table:leaderboard}. Note in Table \ref{table:leaderboard} that although MTR\_E leads to a slight regression in \textit{Distance To Nearest Object Likelihood}, \textit{Collision Likelihood} and \textit{Distance To Road Edge Likelihood}, it significantly improves other component metrics. 

\begin{table*}[t]
	\caption{Performance of our methods on WOSAC Leaderboard.}
	\centering
	\begin{tabular}{c|cccc|c}
 \toprule
		\makecell[c]{Method Name}&\makecell[c]{\textbf{Realism Meta}\\\textbf{metric $\uparrow$}} & \makecell[c]{Kinematic\\metrics $\uparrow$} & \makecell[c]{Interactive\\metrics $\uparrow$} & \makecell[c]{Map-based\\metrics $\uparrow$} & minADE $\downarrow$ \\
		\midrule 
  MTR+++ & 0.4697 & 0.1295 & \textbf{0.1774} & 0.1628  & 1.6817 \\
  MTR\_E & \textbf{0.4911} & \textbf{0.1505} & 0.1766 & \textbf{0.1640} & \textbf{1.6561} \\
		\bottomrule
	\end{tabular}
	\label{table:LB}
\end{table*}

\begin{table*}[t]
\centering
\caption{Detailed performance of our methods on WOSAC Leaderboard.}
 	\begin{tabular}{c|cc}
 \toprule
\makecell[c]{Metric Name} &  \makecell[c]{MTR+++} & \makecell[c]{MTR\_E} \\
		\midrule
 Realism meta-metric & 0.4697 & \textbf{0.4911}\\
 Linear Speed Likelihood & 0.4119 & \textbf{0.4278} \\
 Linear Acceleration Likelihood & 0.1066 & \textbf{0.2353} \\
 Angular Speed Likelihood & 0.4838 & \textbf{0.5335} \\
 Angular Acceleration Likelihood & 0.4365 & \textbf{0.4753} \\
 Distance To Nearest Object Likelihood & \textbf{0.3457} & 0.3455 \\
 Collision Likelihood & \textbf{0.4144} & 0.4091 \\
 Time To Collision Likelihood & 0.7969 & \textbf{0.7983} \\
 Distance To Road Edge Likelihood & \textbf{0.6545} & 0.6541\\
 Offroad Likelihood & 0.5770 & \textbf{0.5840} \\
 minADE & 1.6817 & \textbf{1.6561} \\

		\bottomrule
	\end{tabular}
	\label{table:leaderboard}
\end{table*}

\section{Conclusion}
In this technical report, we have presented the MTR+++ and MTR\_E which simulate multi-agent behaviors by an autoregressive method. In addition, we also propose some simple but effective post-processing algorithms. In future work, we will explore how to better generate diverse and realistic scenarios and how to integrate historical trajectory and generated trajectories in each autoregressive execution. 

\vspace{-0.08in}
\paragraph{Acknowledgements.} 
The authors would like to thank John Lambert for his helpful comments and suggestions that improve this manuscript. 

\vspace{5in}

\bibliographystyle{abbrv}
\bibliography{references}  







\end{document}